\documentclass[runningheads]{llncs}
\usepackage{times}
\usepackage{epsfig}
\usepackage{graphicx}
\usepackage{amsmath}
\usepackage{amssymb}
\usepackage{comment}

\usepackage[utf8]{inputenc}
\usepackage[english]{babel}
\usepackage{booktabs}
\usepackage{adjustbox}
\usepackage[dvipsnames]{xcolor}
\usepackage{subcaption}
\usepackage[ruled]{algorithm2e}
\usepackage{arydshln}
\usepackage{cite}
\usepackage{hyperref}

\DeclareMathOperator*{\argmin}{\arg\!\min}

\def\eg{\emph{e.g}.} 
\def\ie{\emph{i.e}.}

\usepackage[width=122mm,left=12mm,paperwidth=146mm,height=193mm,top=12mm,paperheight=217mm]{geometry}

\begin{document}
\pagestyle{headings}
\mainmatter
\def\ECCVSubNumber{2749}  

\title{End-to-end Interpretable Learning of \\ Non-blind Image Deblurring} 

\begin{comment}
\titlerunning{ECCV-20 submission ID \ECCVSubNumber} 
\authorrunning{ECCV-20 submission ID \ECCVSubNumber} 
\author{Anonymous ECCV submission}
\institute{Paper ID \ECCVSubNumber}
\end{comment}

\titlerunning{End-to-end Interpretable Learning of Non-blind Image Deblurring}
%
\author{Thomas Eboli\inst{1} \and
Jian Sun\inst{2} \and
Jean Ponce\inst{1}}
\authorrunning{T. Eboli et al.}
%
\institute{INRIA, D\'epartement d'informatique de l'ENS, ENS, CNRS, PSL University\\
\email{\{thomas.eboli,jean.ponce\}@inria.fr}\\
\and
Xi'an Jiaotong University\\
\email{jiansun@xjtu.edu.cn}\\
\href{https://github.com/teboli/CPCR}{\textcolor{magenta}{\texttt{https://github.com/teboli/CPCR}}}}

\def\RR{\mathbb{R}}
\def\PP{\mathbb{P}}
\def\AA{\mathbb{A}}
\def\LL{\mathbb{L}}
\def\SS{\mathbb{S}}
\def\barr{\bar{\mathbb{R}}}
\def\mat#1{{\mathcal{#1}}}
\def\vect#1{\mbox{\boldmath $#1$}}
\def\PPi{\mbox{\boldmath$\Pi$}}
\def\squig{\rightsquigarrow}
\def\comment#1{{}}
\def\qmatrix#1{\left[\begin{matrix}#1\end{matrix}\right]}

\renewcommand{\tabcolsep}{2pt}

\maketitle

\begin{abstract}
  Non-blind image deblurring is typically formulated as a linear
  least-squares problem regularized by natural priors on the
  corresponding sharp picture's gradients, which can be solved, for
  example, using a half-quadratic splitting method with Richardson
  fixed-point iterations for its least-squares updates and a proximal operator
   for the auxiliary variable updates. We propose to
  precondition the Richardson solver using approximate inverse filters
  of the (known) blur and natural image prior kernels. 
  Using convolutions instead of a generic linear preconditioner allows
  extremely efficient parameter sharing across the image, and leads to
  significant gains in accuracy and/or speed compared to classical FFT
  and conjugate-gradient methods. More importantly, the proposed architecture
  is easily adapted to learning both the preconditioner and the proximal operator
  using CNN embeddings. This yields
  a simple and efficient algorithm for non-blind image deblurring which is fully interpretable,
  can be learned end to end, and whose accuracy matches or exceeds the
  state of the art, quite significantly, in the non-uniform case.
\end{abstract}


\section{Introduction}

This presentation addresses the problem of non-blind image
deblurring--that is, the recovery of a sharp image given its blurry
version and the corresponding uniform or non-uniform motion blur
kernel. Applications range from photography
\cite{hu16smartphone} to astronomy~\cite{starck06astronomical} and
microscopy~\cite{goodman96introduction}. Classical approaches
to this problem include least-squares and Bayesian models, leading to
Wiener~\cite{wiener49extrapolation} and Lucy-Richardson~\cite{richardson72baysian}
deconvolution techniques for example.  Since many sharp images can
lead to the same blurry one, blur removal is an ill-posed problem.
To tackle this issue, variational
methods~\cite{rudin92totalvariation} inject {\em a priori}
knowledge over the set of solutions using penalized least-squares.
Geman and Yang~\cite{geman1995nonlinear} introduce an auxiliary
variable to solve this problem by iteratively evaluating a proximal
operator~\cite{parikh14proximal} and solving a
least-squares problem. The rest of this presentation builds on this
{\em half-quadratic splitting} approach.  Its proximal part has
received a lot of attention through the design of complex
model-based~\cite{krishnan09hyper, zoran11learning,
  xu13unnatural, sun08image} or learning-based
priors~\cite{roth09field}.  Far less attention had been
paid to the solution of the companion least-squares problem, typically
relying on techniques such as conjugate gradient (CG)
descent~\cite{boyd14convex} or fast Fourier transform
(FFT)~\cite{xu14inverse, hirsch10efficient}. CG is
relatively slow in this context, and it does not exploit the fact that the linear operator
corresponds to a convolution. FFT exploits this
property but is only truly valid under periodic conditions at the
boundaries, which are never respected by real images.

We propose instead to use Richardson fixed-point iterations \cite{kelley95iterative} to solve
the least-squares problem, using approximate inverse filters of the
(known) blur and natural image prior kernels as preconditioners.
Using convolutions instead of a traditional linear preconditioner allows efficient parameter sharing across the image, which leads to
significant gains in accuracy and/or speed over FFT
and conjugate-gradient methods.
To further improve
performance and leverage recent progress in
deep learning, several recent approaches to denoising and deblurring
unroll a finite number of proximal updates and least-squares
minimization steps~\cite{schmidt14shrinkage,
chen17trainable, kruse17learning,
aljadaany19douglas, kobler17variational}. 
Compared to traditional convolutional neural networks (CNNs), these algorithms use interpretable components and produce intermediate feature maps that can be directly supervised during training \cite{schmidt14shrinkage, kruse17learning}.

We propose a solver for non-blind deblurring, also based on the splitting scheme of~\cite{geman1995nonlinear} but, in addition to
learning the proximal operator as
in~\cite{zhang17fcnn}, we also learn parameters in the fixed-point algorithm by embedding the preconditioner into a CNN whose bottom layer's kernels are the approximate filters discussed above. 
Unlike the algorithm of \cite{zhang17fcnn}, our algorithm is trainable
end to end, and achieves accuracy that matches or exceeds the state of the art.
Furthermore, in contrast to other state-of-the-art CNN-based methods \cite{zhang17fcnn, kruse17learning} relying on FFT, it operates in the pixel domain and thus easily extends to non-uniform blurs scenarios.

\subsection{Related work}

\textbf{Uniform image deblurring.}~Classical priors for natural images minimize the magnitude of gradients using the $\ell_2$~\cite{rudin92totalvariation}, $\ell_1$~\cite{levin09understanding} and hyper-Laplacian~\cite{krishnan09hyper} (semi) norms or parametric potentials~\cite{sun08image}.
Instead of restoring the whole image at once, some works focus on patches by learning their probability distribution~\cite{zoran11learning} or exploiting local properties in blurry images~\cite{michaeli14blind}.
{Handcrafted} priors are designed so that the optimization is feasible and easy to carry out but they may ignore the characteristics of the images available.
Data-driven priors, on the other hand, can be learned from a training dataset  (\textit{e.g.,} new regularizers).
Roth and Black~\cite{roth09field} introduce a learnable total variation (TV) prior whose parameters are potential/filter pairs.
This idea has been extended in shallow neural networks in \cite{schmidt13discriminative, schmidt14shrinkage} based on the splitting scheme of \cite{geman1995nonlinear}.
Deeper models based on Roth and Black's learnable prior had been proposed ever since \cite{chen17trainable, kobler17variational, kruse17learning}.
The proximal operator can also be replaced by a CNN-based denoiser~\cite{zhang2017learning, meinhardt17learning} or a CNN specifically trained to mimic a proximal operator~\cite{zhang17fcnn} or the gradient of the regularized~\cite{gong2020deepgradient}.
More generally, CNNs are now used in various image restoration tasks, including non-blind deblurring by refining a low-rank decomposition of a deconvolution filter kernel \cite{xu14deep}, using a FFT-based solver \cite{schuler16learning}, or to improve the accuracy of splitting techniques \cite{aljadaany19douglas}. 

\textbf{Non-uniform image deblurring.}~Non-uniform deblurring is more challenging than its uniform counterpart \cite{cho07removing, chakrabarti10analyzing}. 
Hirsch \text{et al.}~\cite{hirsch10efficient} consider large overlapping patches and suppose uniform motion on their supports before removing the local blur with an FFT-based uniform deconvolution method.
Other works consider pixelwise locally linear motions~\cite{sun15learning, kim14segmentation, brooks19synthesize, couzinie13learning} as simple elements representing complex global motions and solve penalized least-squares problems to restore the image.
Finally, geometric non-uniform blur can be used in the case of camera shake to predict motion paths~\cite{whyte12nonuniform, tai11richardson}.

\subsection{Main contributions}

Our contributions can be summarized as follows.
\begin{itemize}
    \item We introduce a convolutional preconditioner for fixed-point
      iterations that efficiently solves the least-squares problem
      arising in splitting algorithms to minimize
      penalized energies.  It is faster and/or more
      accurate than FFT and CG for this task, with theoretical convergence guarantees.
    \item We propose a new end-to-end trainable algorithm that
      implements a finite number of stages of half-quadratic splitting \cite{geman1995nonlinear} and is fully interpretable.
      It alternates between proximal updates and preconditioned
      fixed-point iterations.  
      The proximal operator and linear preconditioner are parameterized by CNNs in order to learn these functions from a training set of clean and blurry
      images.
    \item We evaluate our approach on several benchmarks with both uniform and non-uniform blur kernels.
    We demonstrate its robustness to significant levels of noise, and obtain results that are competitive with the state of the art for uniform blur and significantly outperforms it in the non-uniform case.
\end{itemize}

\section{Proposed method}
Let $y$ and $k$ respectively denote a blurry image and a known
blur kernel. The {\em deconvolution} (or {\em non-blind deblurring})
problem can be formulated as
\begin{equation}
  \min_{x} \frac{1}{2}|| y- k\star x||_F^2+\lambda \Omega ( \sum_{i=1}^n k_i \star x ),
  \label{eq:penalizedproblem}
\end{equation}
where ``$\star$'' is the convolution operator, and
$x$ is the (unknown) sharp image. 
The filters $k_i$ ($i=1,
\ldots, n$) are typically partial derivative operators, and $\Omega$ acts as a
regularizer on $x$, enforcing natural image priors. One
often takes $\Omega(z)~=~||z||_1$ (TV-$\ell_1$ model).
We propose in this section an end-to-end learnable variant of the method of {\em half-quadratic splitting} (or {\em HQS}) \cite{geman1995nonlinear} to
solve Eq.~(\ref{eq:penalizedproblem}).  As shown later, a key to the
effectiveness of our algorithm is that all linear operations are explicitly
represented by convolutions. 

Let us first introduce notations that will simplify the presentation.
Given some linear filters $a_i$ and $b_i$ ($i=0,\ldots,n$) with
finite support (square matrices), we borrow the Matlab notation for ``stacked''
linear operators, and denote by $A=[a_0,\ldots,a_n]$ and
$B=[b_0;\ldots;b_n]$ the (convolution) operators respectively obtained
by stacking ``horizontally'' and ``vertically'' these filters, whose responses are
\begin{equation}
	\begin{array}{l}
  A\star x =[a_0\star x,\ldots,a_n\star x];\;
  B\star x =[(b_0\star x)^\top,\ldots,(b_n\star x)^\top]^\top;\;
  \end{array}
\end{equation}
We also define $A\star B=\sum_{i=0}^n a_i\star
b_i$ and easily verify that $(A\star B)\star x=A\star(B\star
x)$.
\subsection{A convolutional HQS algorithm}
\label{sec:convolutionalhqsalgorithm}
Equation~(\ref{eq:penalizedproblem}) can be rewritten as
\begin{equation}
  \min_{x,z} \frac{1}{2}||y-k\star x||_F^2+\lambda \Omega(z)
 \;\text{such that}\; z=F\star x,
\end{equation}
where $F=[k_1;\ldots;k_n]$.
Let us define the energy function
\begin{equation}
  E(x,z,\mu)=\frac{1}{2}||y-k\star x||_F^2+\lambda\Omega(z)
    +\frac{\mu}{2}||z-F\star x||_F^2.
\end{equation}
Given some initial guess $x$ for the sharp image, (\eg~$x~=~y$) we can
now solve our original problem using the {\em HQS} method \cite{geman1995nonlinear} with $T$ iterations of the
form
\begin{equation}
  \begin{array}{l}
    z\leftarrow\argmin_{z} E (x, z,\mu);\\
    x\leftarrow\argmin_{x} E (x,z,\mu);\\
    \mu\leftarrow\mu+\delta t.
  \end{array}
\end{equation}
The $\mu$ update can vary with iterations but must be positive. We could also use the alternating direction method of multipliers (or {\em ADMM} \cite{parikh14proximal}), for example, but this is left to future work.
Note that the update in $z$ has the form
\begin{equation}\label{eq:proximalzupdate}
  z\leftarrow\argmin_{z} \frac{\mu}{2}||z-F\star x||_F^2
  +\lambda\Omega(z) =\varphi_{\lambda/\mu}(F\star x),
\end{equation}
where $\varphi_{\lambda/\mu}$ is, by definition, the {\em proximal operator} \cite{parikh14proximal} associated with
$\Omega$ (a soft-thresholding function in the case of the $\ell_1$ norm \cite{elad10sparse})
given $\lambda$ and $\mu$.

The update in $x$ can be written as the solution of a linear
least-squares problem:
\begin{equation}
  x\leftarrow\argmin_{x} \frac{1}{2}||u-L\star x||_F^2,
  \label{eq:xupdate}
\end{equation}
where $u=[y;\sqrt{\mu} z]$ and $L=[k;\sqrt{\mu}F]$.

\subsection{Convolutional PCR iterations}
Many methods are of course available for solving Eq.~(\ref{eq:xupdate}).
We propose to compute $x$ as the solution of $C \star
(u - L \star x)=0$, where $C=[c_0,\ldots,c_n]$ is
composed of $n+1$ filters 
and is used in {\em preconditioned Richardson} (or {\em PCR}) fixed-point iterations~\cite{kelley95iterative}.

Briefly, in the generic linear case, PCR is an iterative method for
solving a square, nonsingular system of linear equations
$A x=b$. Given some initial estimate $x=x_0$ of the unknown $x$, it repeatedly applies the iterations
\begin{equation}
  x\leftarrow x - C ( A x - b),
\end{equation}\label{eq:preconditionedxupdate}
where $C$ is a preconditioning square matrix. When $C$ is an {\em
approximate inverse} of $A$, that is, when the spectral radius
$\eta$ of $\text{Id}- C A$ is smaller than one, preconditioned
Richardson iterations converge to the solution of $A x = b$ with a linear
rate proportional to $\eta$~\cite{kelley95iterative}. When $A$ is an $m\times
n$ matrix with $m\ge n$, and $x$ and $b$ are respectively elements of
$\RR^n$ and $\RR^m$, PCR can also be used in Cimmino's algorithm for
linear least-squares, where the solution of $\min_{x}
||A x - b||^2$ is found using $C=\rho A^\top$, with $\rho>0$ sufficiently
small, as the solution of $A^\top A x-A^\top b=0$, with
similar guarantees. Finally, it is also possible to use a different $n\times m$
matrix. When the spectral radius $\eta$ of the $n\times n$
matrix $\text{Id}- C A$ is smaller than one, the PCR iterations converge
once again at a linear rate proportional to $\eta$.  However, they converge to the (unique in general) solution of
$C( A x- b)=0$, which may of course be different from the least-squares solution.

 \begin{figure}[t]
     \adjustbox{max width=0.99\textwidth}{
    \centering
    \begin{tabular}{ccc}
        \begin{subfigure}[b]
        {0.33\textwidth}
        \centering\includegraphics[scale=0.11]{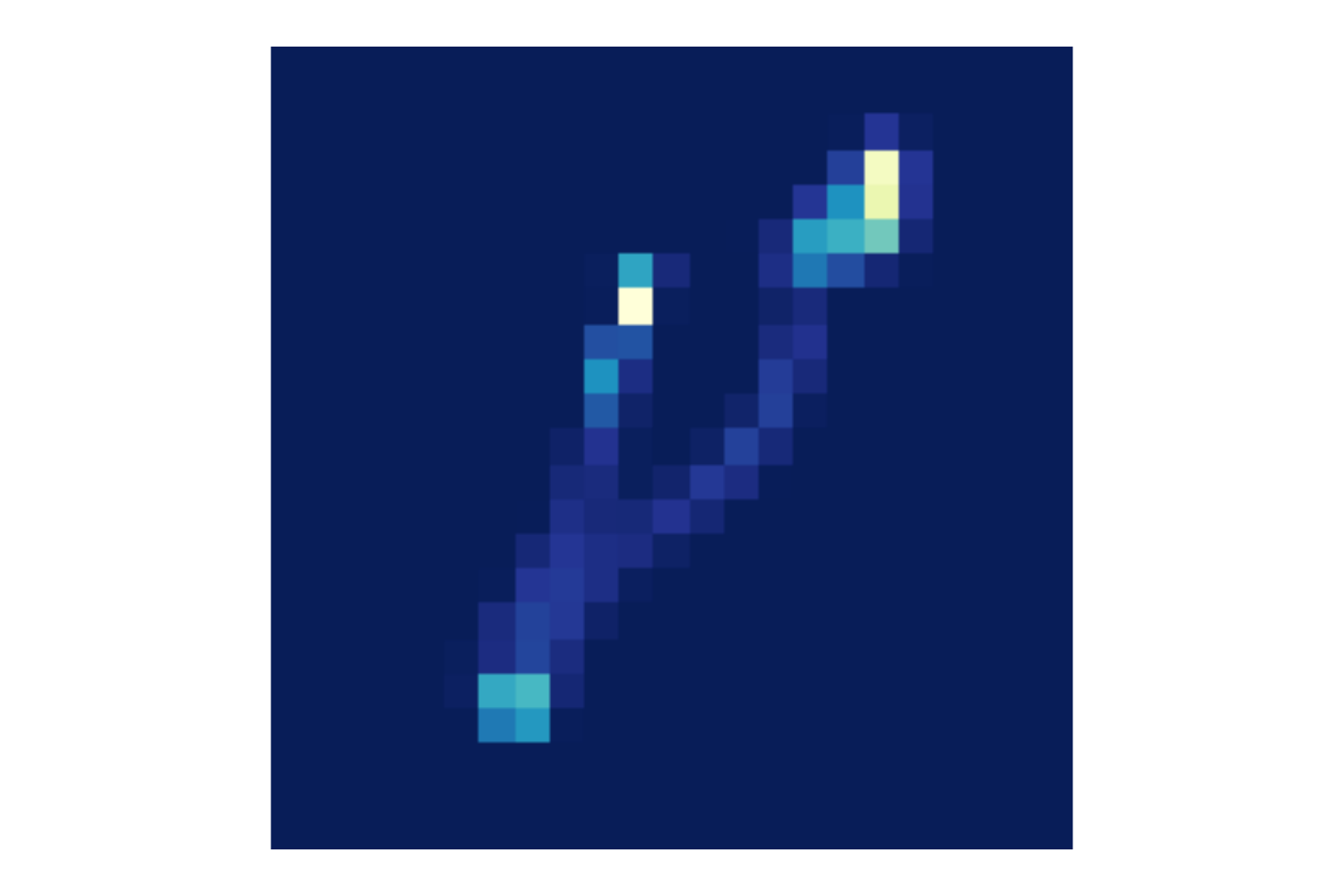}\label{fig:k}\end{subfigure} & 
        \begin{subfigure}[b]
        {0.33\textwidth}
        \centering\includegraphics[scale=0.24]{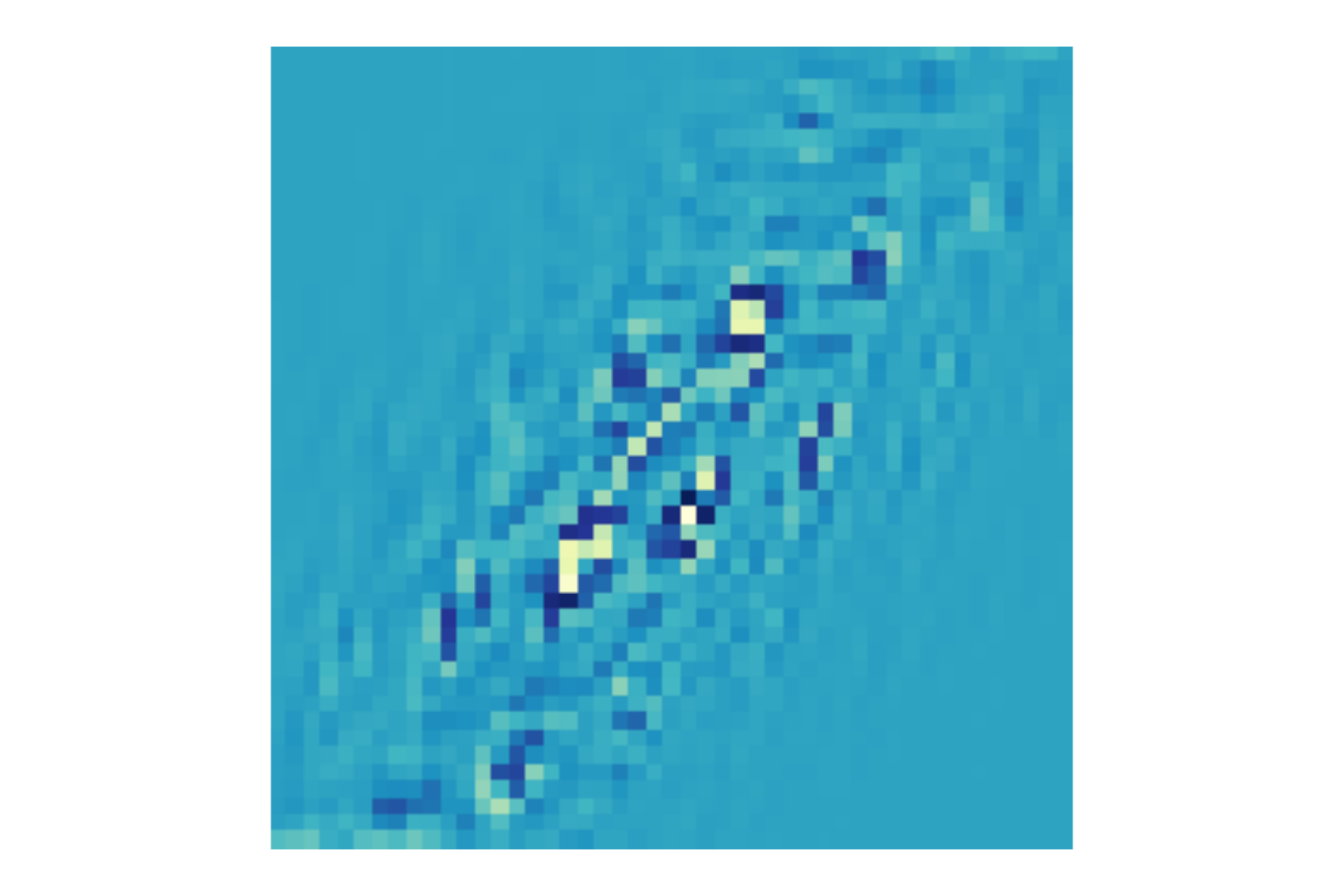}\label{fig:c}\end{subfigure} & 
        \begin{subfigure}[b]
        {0.33\textwidth}
        \centering\includegraphics[scale=0.20]{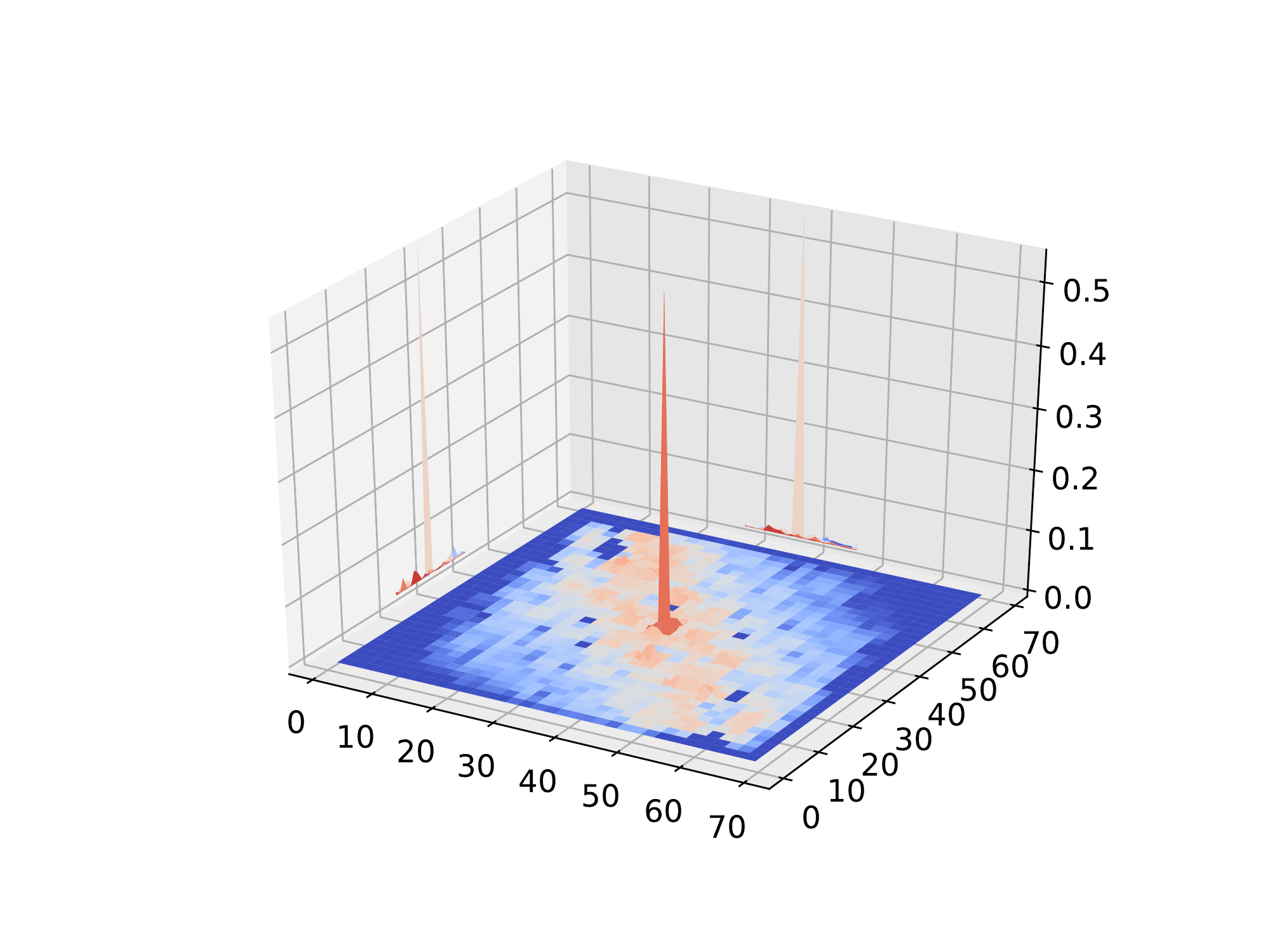}\label{fig:d}\end{subfigure} \\
    \end{tabular}
    }
    \caption{From left ot right: An example of a blur kernel $k$ from the Levin \textit{et al.} dataset \cite{levin09understanding}; its approximate inverse kernel $c_0$; the resulting filter resulting from the convolution of $k$ and $c_0$ (represented as a surface). It gives an approximate Dirac filter $\delta$.}
    \label{fig:exampleofkernels}
\end{figure}

This method is easily adapted to our context.
Since $L$ corresponds to a bank of filters of size $w_k \times w_k$, it is natural to take
$C=[c_0,\ldots,c_n]$ to be another bank of $n~+~1$ linear filters of size $w_c \times w_c$.
Unlike a generic linear preconditioner satisfying $CA \approx$~Id in matrix form, whose size
depends on the square of the image size, $C$ exploits the structure of $L$ and
is a linear operator with {\em much} fewer parameters, \ie~$n~+~1$ times the size of the $c_i$'s.
Thus, $C$ is an approximate inverse filter bank for
$L$, in the sense that
\begin{equation}
  \textstyle \delta \approx
  L \star C = C \star L = c_0\star k + \sqrt{\mu} \sum_{i=1}^n c_i \star k_i,
\label{eq:invfil}
\end{equation}
where $\delta$ is the Dirac filter. 
 In this setting, $C$ is computed
as the solution of
\begin{equation}
  \textstyle C =  \argmin_C ||\delta - L \star C||_F^2+\rho
  \sum_{i=0}^n || c_i ||_F^2,
\label{eq:ridge}
\end{equation}
The classical solution using the pseudo inverse of $L$ has cost $\mathcal{O}\left((w_k + w_c - 1)^{2\times 3}\right)$.
\begin{equation}\label{eq:fftinversekernel}
    c_i = \mathcal{F}^{-1}\left(\frac{\tilde K_i^*}{\rho J + \sum_{j=0}^n | \tilde K_j |^2}\right) \quad \text{for } i = 0\text{ to }n, 
\end{equation}
using the fast Fourier transform (FFT) with cost $\mathcal{O}(w_c^2\log (w_c))$ \cite{folland92fourier}. $\mathcal{F}^{-1}$ is the inverse Fourier transform, $J$ is a matrix full of ones, $\tilde K_i$ is the Fourier transform of $k_i$ (with $k_0=k$), $\tilde K_i^*$ is its complex conjugate and the division in the Fourier domain is entrywise. 
Note that the use of FFT in this context has nothing to do with its use as a deconvolution tool for solving Eq.~\eqref{eq:xupdate}.
Figure~\ref{fig:exampleofkernels} shows an example of a blur kernel from \cite{levin09understanding}, its approximate inverse when $n=0$ and the result of their convolution.
Let us define $[A]_{\star}$ as the linear operator such that $[A]_{\star}B = A \star B$.
Indeed, a {\em true}
inverse filter bank such that equality holds in Eq.~(\ref{eq:invfil})
does not exist in general (\textit{e.g.,} a Gaussian filter cannot be inverted),
but all that matters is that the linear operator associated with
$\delta-C\star L$ has a spectral radius smaller than one~\cite{kelley95iterative}.
We have the following result.

\begin{lemma}
The spectral radius of the linear operator $\text{Id}-[L]_{\star}[C]_{\star}$, where $C$ is the optimal solution of \eqref{eq:ridge} given by \eqref{eq:fftinversekernel} is always smaller than $1$ when $[L]_{\star}$ has full rank.
\end{lemma}
A detailed proof can be found in the supplemental material. 
We now have our basic non-blind deblurring algorithm, in the form of
the Matlab-style CHQS (for {\em convolutional HQS}, primary) and CPCR
(for {\em convolutional PCR}, auxiliary) functions below.

{\small
\begin{center}
    \begin{minipage}{0.62\textwidth}
        \begin{center}
        \fbox{
        \parbox[c][110pt]{0.49\textwidth}{
        \begin{tabbing}
            \textbf{function} $x=\text{CHQS}(y, k, F, \mu0)$\\
                $x=y$; $\mu=\mu0$; \\
            \textbf{for} \=
                $t=0:T-1$ \textbf{do}\\
            \> $u=[y;\sqrt{\mu}\varphi_{\lambda/\mu}(F \star x)]$;\\
            \> $L=[k;\sqrt{\mu} F]$;\\
            \> $C=\argmin_{C} ||\delta-C \star L||_F^2+\rho
            \sum_{i=0}^n|| c_i||_F^2$;\\
            \> $x=\text{CPCR}(L,u,C,x)$;\\ \>
            $\mu=\mu+\delta_t;$\\ 
            \textbf{end for}\\
            \textbf{end function}
        \end{tabbing} 
        }   
        }
        \end{center}
    \end{minipage}%
    \begin{minipage}{0.38\textwidth}
        \begin{center}
        \fbox{
        \parbox[c][68pt]{1.0\columnwidth}{
        \begin{tabbing}
          \textbf{function} $x=\text{CPCR}(A,b,C,x_0)$  \\
          $x = x_0;$\\
          \textbf{for} \= $s=0:S-1$ \textbf{do}\\
          \> $x=x-C\star(A\star x- b)$;\\
           \textbf{end for}\\
          \textbf{end function}
        \end{tabbing}
        }
        }
        \end{center}
    \end{minipage}
\end{center}
}

\subsection{An end-to-end trainable CHQS algorithm}
To improve on this method, we propose to learn the proximal operator
$\varphi$ and the preconditioning operator $C$. 
The corresponding
{\em learnable} CHQS (LCHQS) algorithm can now be written as a function
with two additional parameters $\theta$ and $\nu$ as follows.

{\small 
\begin{center}
\fbox{
\parbox[c][110pt]{0.49\textwidth}{
\begin{tabbing}
  \textbf{function} $x=\text{LCHQS}(y,k,F,\mu_0,\theta,\nu)$\\
  $x=y$; $\mu=\mu_0$; \\
  \textbf{for} \=
  $t=0:T-1$ \textbf{do}\\
  \> $u=[y;\sqrt{\mu}\varphi_{\lambda/\mu}^{\theta}(F\star\ x)]$;\\
  \> $L=[k;\sqrt{\mu}F]$;\\
  \> $C=\argmin_{C} ||\delta - C \star L||_F^2+\rho
    \sum_{i=0}^n|| c_i||_F^2$;\\
  \> $x=\text{CPCR}(L,u,\psi^{\nu}(C),x)$;\\ \>
  $\mu=\mu+\delta_t;$\\ 
   \textbf{end for}\\
 \textbf{end function}
\end{tabbing}
}
}
\end{center}
}

The function LCHQS has the same structure as CHQS but now 
uses two parameterized embedding functions $\varphi_{\tau}^{\theta}$ and $\psi^{\nu}$ for the proximal operator and preconditioner.
In practice, these functions are CNNs with learnable parameters $\theta$ and $\nu$ as detailed in Sec.~\ref{sec:Experiments}.
Note that $\theta$ actually determines the regularizer through its proximal operator.
The function LCHQS is differentiable with respect to both its $\theta$
and $\nu$ parameters.  
Given a set of training triplets $(x^{(i)}, y^{(i)}, k^{(i)})$
(in $i=1,\ldots,N$), the parameters $\theta$ and $\nu$ can thus be
learned end-to-end by minimizing
\begin{equation}\label{eq:objectivefull}
    \textstyle F(\theta,\nu)=\sum_{i=1}^N ||x^{(i)}-\text{LCHQS}(y^{(i)}, k^{(i)},F,\theta,\nu)||_1,
\end{equation}
with respect to these two parameters by ``unrolling'' the HQS iterations
and using backpropagation, as in \cite{zhang17fcnn, chen17trainable} for example.
This can be thought of as the ``compilation'' of a fully interpretable iterative optimization algorithm into a CNN architecture.
Empirically, we have found that the $\ell_1$ norm gives better results than the $\ell_2$ norm in Eq.~\eqref{eq:objectivefull}.

\section{Implementation and results}
\label{sec:Experiments}

\begin{figure}[t]
	\centering
	\adjustbox{max width=0.99\textwidth}{
	\begin{tabular}{ccc}
		\begin{subfigure}[t]{0.33\textwidth}\centering\includegraphics[scale=0.23]{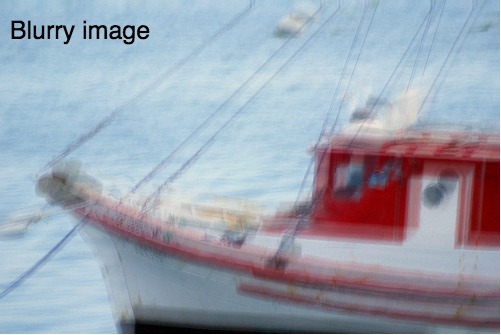}\end{subfigure} & 
		\begin{subfigure}[t]{0.33\textwidth}\centering\includegraphics[scale=0.23]{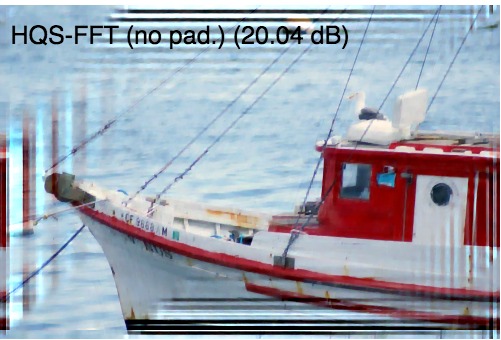}\end{subfigure} & 
		\begin{subfigure}[t]{0.33\textwidth}\centering\includegraphics[scale=0.23]{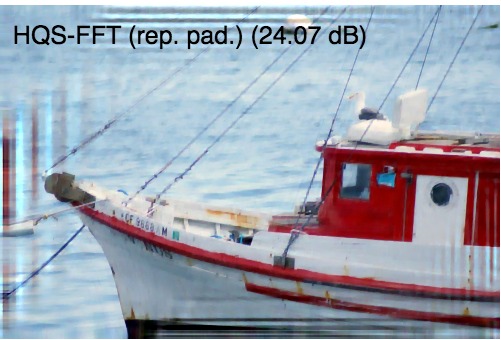}\end{subfigure} \\
		\begin{subfigure}[t]{0.33\textwidth}\centering\includegraphics[scale=0.23]{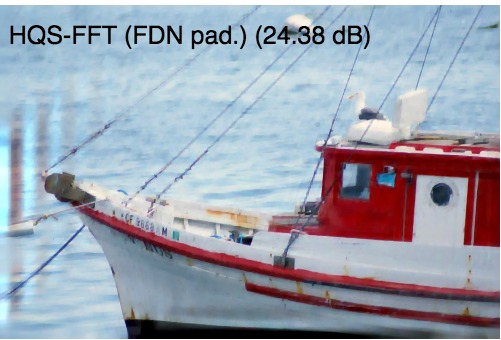}\end{subfigure} & 
		\begin{subfigure}[t]{0.33\textwidth}\centering\includegraphics[scale=0.23]{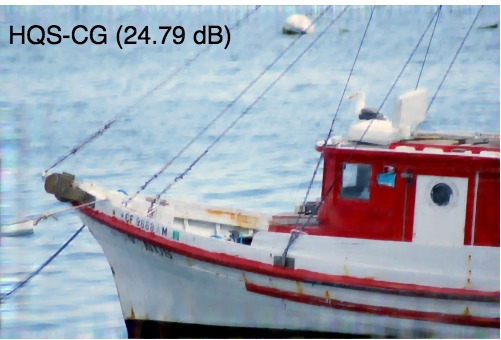}\end{subfigure} & 
		\begin{subfigure}[t]{0.33\textwidth}\centering\includegraphics[scale=0.23]{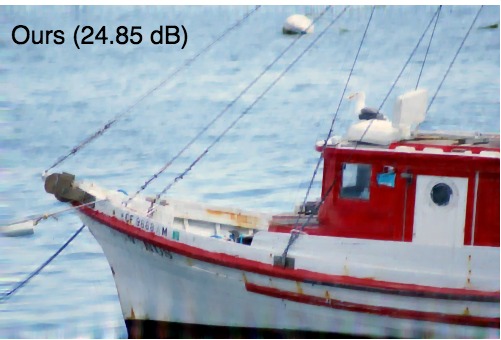}\end{subfigure} \\
	\end{tabular}
	}
	\caption{A blurry image from our test set with $2\%$ white noise and the 
	solutions of Eq.~\eqref{eq:penalizedproblem} with TV-$\ell_1$ regularization obtained with different HQS-based methods. 
	From the same optimization problem, HQS-FFT displays boundary artifacts. HQS-CG and CHQS produce images with similar visual quality and PSNR values but HQS-CG is much slower.}
	\label{fig:qualitativetvl1model}
\end{figure}

\subsection{Implementation details}

\textbf{Network architectures.}~The global architecture of LCHQS shares the same pattern than FCNN \cite{zhang17fcnn}, \ie,~$n=2$ in Eq.~\eqref{eq:penalizedproblem} with $k_1=[1,-1]$ and $k_2=k_1^\top$, and the model repeats between 1 and 5 stages alternatively solving the proximal problem  \eqref{eq:proximalzupdate} and the linear least-squares problem \eqref{eq:xupdate}.
The proximal operator $\varphi^{\theta}$ is the same as the one introduced in \cite{zhang17fcnn}, and it is composed of 6 convolutional layers with 32 channels and $3 \times 3$ kernels, followed by ReLU non-linearities, except for the last one.
The first layer has 1 input channel and the last layer has 1 output channel.
The network $\psi^{\nu}$ featured in LCHQS is composed of 6 convolutional layers with 32 channels and $3 \times 3$ kernels, followed by ReLU non-linearities, except for the last one.
The first layer has $n+1$ input channels (3 in practice with the setting detailed above) corresponding to the filtered versions of $x$ with the $c_i$'s, and the last layer has 1 output channel. 
The filters $c_1$ and $c_2$ are of size $31 \times 31$.
This size is intentionally made relatively large compared to the sizes of $k_1$ and $k_2$ because inverse filters might have infinite support in principle.
The size of $c_0$ is twice the size of the blur kernel $k$.
This choice will be explained in Sec.~\ref{sec:experimentalvalidationofcpcrandchqs}.
In our implementation, each LCHQS stage has its own $\theta$ and $\nu$ parameters.
The non-learnable CHQS module solves a TV-$\ell_1$ problem; the proximal step implements the soft-thresholding operation $\varphi$ with parameter $\lambda / \mu$ and the least-squares step implements CPCR. The choice of $\mu$ will be detailed below.

\textbf{Datasets.}~The training set for uniform blur is made of 3000 patches of size $180 \times 180$ taken from BSD500 dataset and as many random $41 \times 41$ blur kernels synthesized with the code of~\cite{chakrabarti16neural}.
We compute ahead of time the corresponding inverse filters $c_i$ and set the size of $c_0$ to be $83 \times 83$ with Eq.~\eqref{eq:fftinversekernel} where $\rho$ is set to 0.05, a value we have chosen after cross-validation on a separate test set.
We also create a training set for non-uniform motion blur removal made of 3000 $180 \times 180$ images synthesized with the code of~\cite{gong17motion} with a locally linear motion of maximal magnitude of 35 pixels. 
For both training sets, the validation sets are made of 600 additional samples.
In both cases, we add Gaussian noise with standard deviation matching that of the test data.
We randomly flip and rotate by $90^{\circ}$ the training samples and take $170 \times 170$ random crops for data augmentation.

\textbf{Optimization.}~Following \cite{kruse17learning}, we train our model in a two-step fashion: 
First, we supervise the sharp estimate output by each iteration of LCHQS in the manner of \cite{kruse17learning} with Eq.~\eqref{eq:objectivefull}. 
We use an Adam optimizer with learning rate of $10^{-4}$ and batch size of 1 for 200 epochs. 
Second, we further train the network by supervising the final output of LCHQS with Eq.~\eqref{eq:objectivefull} on the same training dataset with an Adam optimizer and learning rate set to $10^{-5}$ for 100 more epochs {\em without} the per-layer supervision.
We have obtained better results with this setting than using either of the two steps separately.

\begin{figure*}[t]
    \centering
    \adjustbox{max width=0.99\textwidth}{
	\begin{tabular}{ccc}
	    \begin{subfigure}[t]{0.33\textwidth}\centering\includegraphics[scale=0.145]{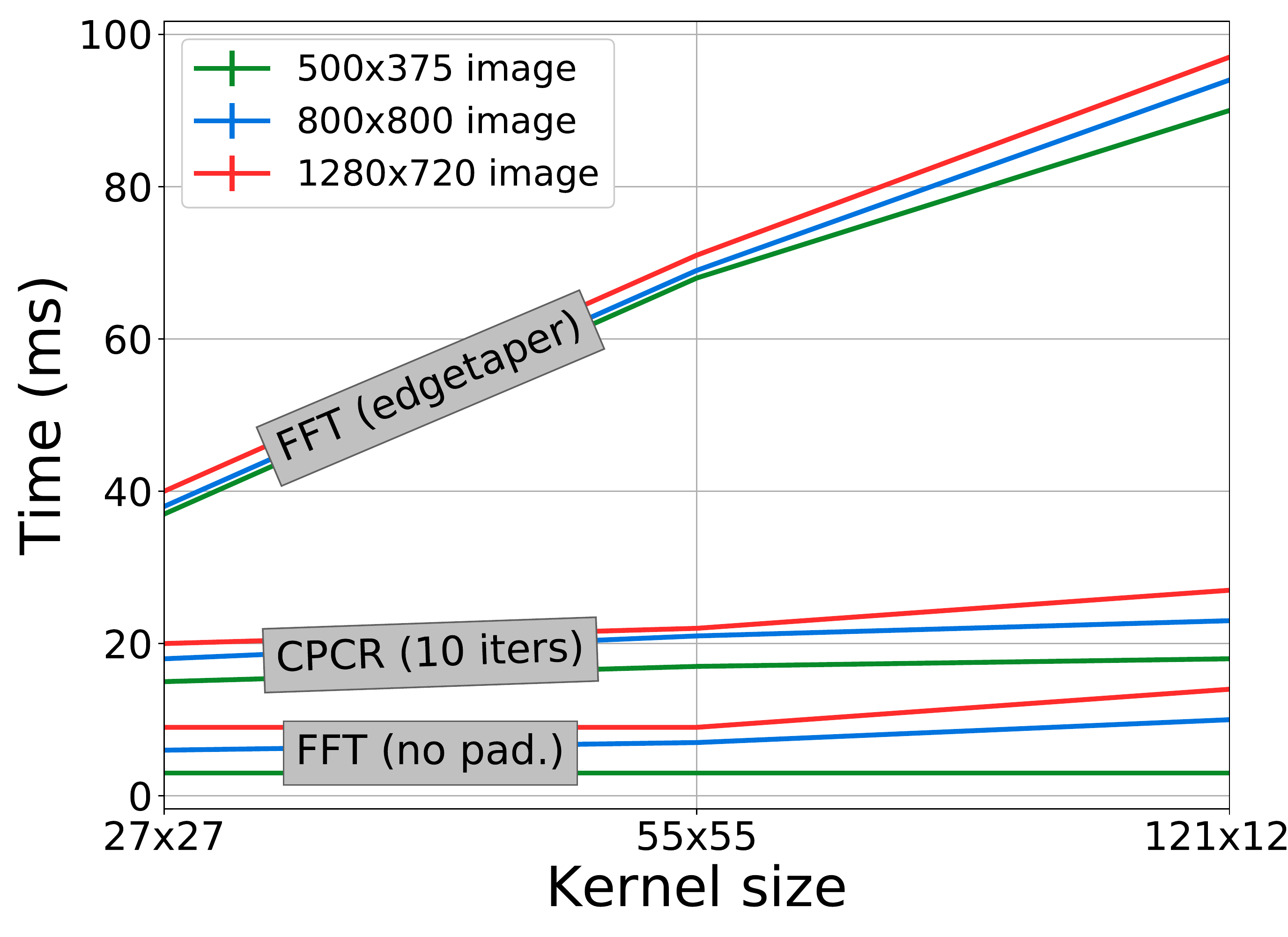}
	    \label{fig:cpcrspeed}\end{subfigure} &
	    \begin{subfigure}[t]{0.33\textwidth}\centering\includegraphics[scale=0.145]{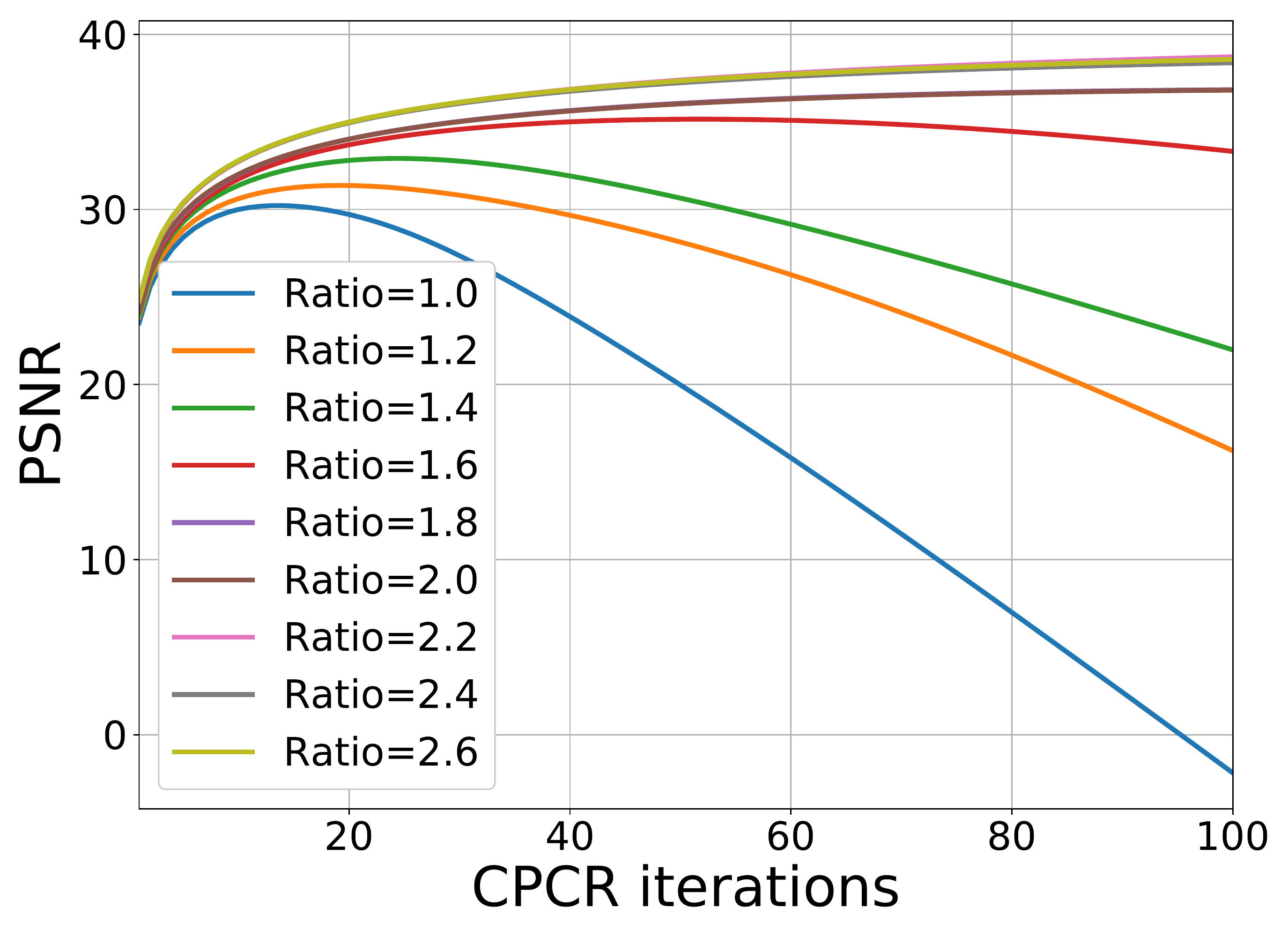}
	    \label{fig:comparescalesirc}\end{subfigure} &
	    \begin{subfigure}[t]{0.33\textwidth}\centering\includegraphics[scale=0.145]{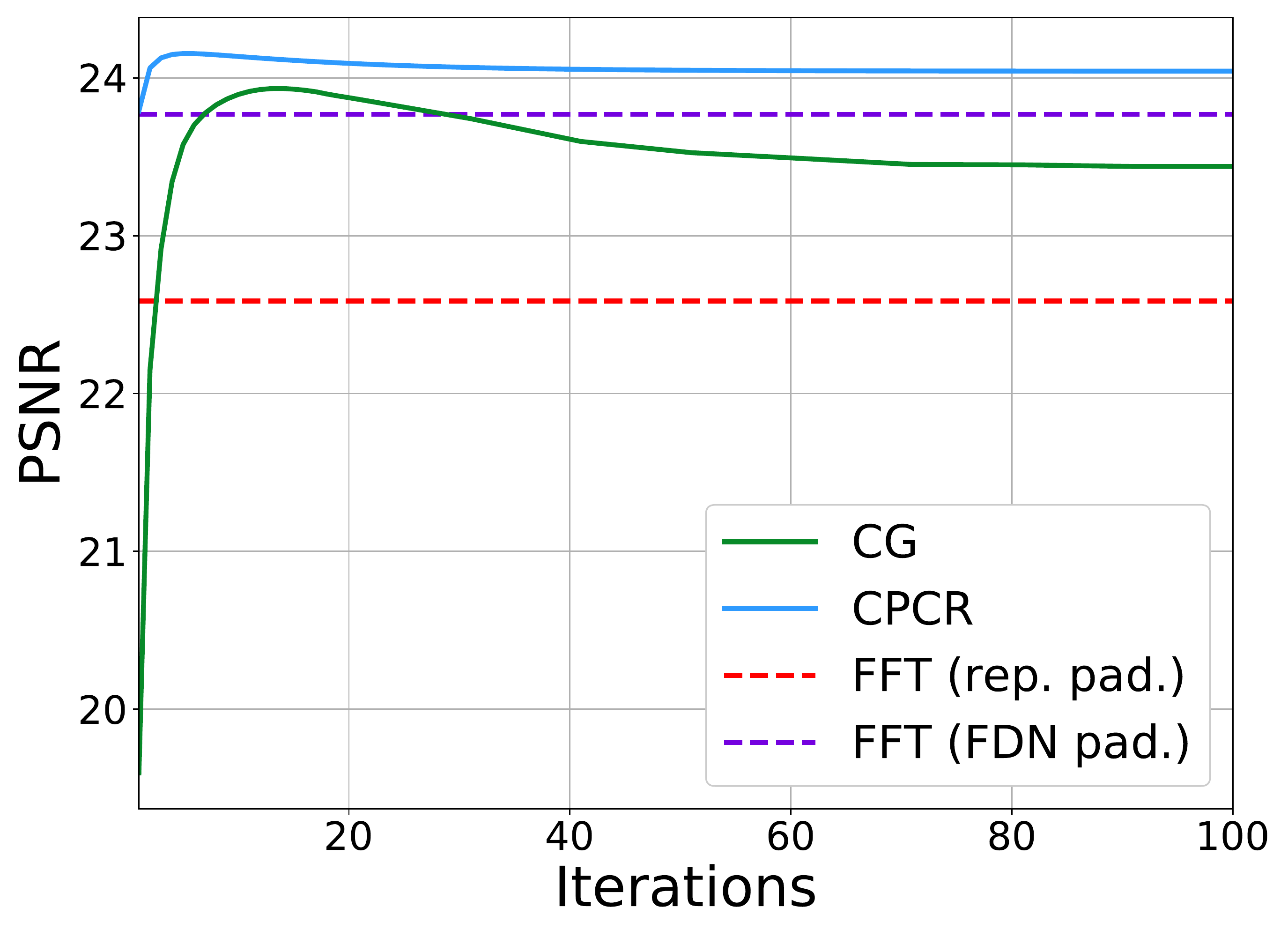}
	   
	    \label{fig:compareoneiterationtvl1curves}\end{subfigure} \\
	\end{tabular}
	}
	\caption{From left to right: Computation times for CPCR (including computation of $C$) with FFT applied on images padded with ``edgetaper'' as recommended in \cite{kruse17learning} and non-padded images for three image formats; effect of the $w_{c_0}/w_{k}$ ratio on performance; comparison of CG, FFT and CPCR for solving \eqref{eq:xupdate}.}
	\label{fig:validations}
\end{figure*}

\subsection{Experimental validation of CPCR and CHQS}
\label{sec:experimentalvalidationofcpcrandchqs}

In this section, we present an experimental sanity check of CPCR for solving \eqref{eq:xupdate} and CHQS for solving \eqref{eq:penalizedproblem} in the context of a basic TV-$\ell_1$ problem.

\textbf{Inverse kernel size.}~We test different sizes for the $w_{c_0} \times w_{c_0}$ approximate inverse filter $c_0$ associated with a $w_{k} \times w_{k}$ blur kernel $k$,
in the non-penalized case, with $\lambda~=~0$.
We use Eq.~\eqref{eq:fftinversekernel} with $\rho$ set to 0.05.
We use 160 images obtained by applying the 8 kernels of~\cite{levin09understanding} to 20 images from the Pascal VOC 2012 dataset. As shown in Fig.~\ref{fig:validations}
, the PSNR increases with increasing $w_{c_0}/w_{k}$ ratios, but saturates when the ratio is larger than 2.2. 
We use a ratio of 2 which is a good compromise between accuracy and speed.

\textbf{CPCR accuracy.}
We compare the proposed CPCR method to FFT-based deconvolution (FFT) and conjugate gradient descent (CG), to solve the least-squares problem of Eq.~\eqref{eq:xupdate} in the setting of a TV-$\ell_1$ problem.
We follow \cite{kruse17learning} and, in order to limit boundary artifacts for FFT, we pad the images to be restored by replicating the pixels on the boundary with a margin of half the size of the blur kernel and then use the ``edgetaper'' routine.
We also run FFT on images padded with the ``replicate'' strategy consisting in simply replicating the pixels on the boundary.
We solve Eq.~\eqref{eq:xupdate} with $\mu_0 = 0.008$, $\lambda = 0.003$ and $z$ computed beforehand with Eq.~\eqref{eq:proximalzupdate}.
The 160 images previously synthesized are degraded with $2\%$ additional white noise.
Figure~\ref{fig:validations}
shows the average PSNR scores for the three algorithms optimizing Eq.~\eqref{eq:xupdate}.
After only 5 iterations, CPCR produces an average PSNR higher than the other methods and converges after 10 iterations.
The ``edgetaper'' padding is crucial for FFT to compete with CG and CPCR by reducing the amount of border artifacts in the solution.

\begin{table}[t]
\centering
\adjustbox{max width=0.99\textwidth}{
	\begin{tabular}{@{}lcccccccccc@{}} 
		\toprule
		 & ker-1 & ker-2 & ker-3 & ker-4 & ker-5 & ker-6 & ker-7 & ker-8 & Aver. & Time (s)\\
		\midrule
		HQS-FFT (no pad.) & 21.14 & 20.51 & 22.31 & 18.21 & 23.36 & 20.01 & 19.93 & 19.02 & 20.69 & 0.07 \\	
		HQS-FFT (rep. pad.) & \underline{26.45} & 25.39 & \textbf{26.27} & 22.75 & 27.64 & 27.26 & 24.84 & 23.54 & 25.53 & 0.07 \\	
		HQS-FFT (FDN pad.) & \textbf{26.48} & 25.89 & \textbf{26.27} & 23.79 & \underline{27.66} & 27.23 & 25.26 & 25.02 & 25.96 & 0.15 \\	

		HQS-CG  & 26.39 & \underline{25.90} & 26.24 & \underline{24.88} & 27.59 & \underline{27.31} & \underline{25.39} & \underline{25.19} & \underline{26.12} & 13 \\
		\hdashline
		CHQS & \underline{26.45} & \textbf{25.96} & \underline{26.26} & \textbf{25.06} & \textbf{27.67} & \textbf{27.51} & \textbf{25.81} & \textbf{25.48} & \textbf{26.27} & 0.26 \\
		\bottomrule	
	\end{tabular}
	}
    \caption{Comparison of different methods optimizing the same TV-$\ell_1$ deconvolution model \eqref{eq:penalizedproblem} 
    on 160 synthetic blurry images with $2\%$ white noise.
    We run all the methods on a GPU. The running times are for a $500 \times 375$ RGB image.}
	\label{tab:comparedtvl1model}
\end{table}

\textbf{CPCR running time.}~CPCR relies on convolutions and thus greatly benefits from GPU acceleration.
For instance, for small images of size $500 \times 375$ and a blur kernel of size $55 \times 55$, 10 iterations of CPCR are in the ballpark of FFT without padding: CPCR runs in 20ms, FFT runs in 3ms and FFT with ``edgetaper'' padding takes 40ms.
For a high-resolution 1280 $\times$ 720 image and the same blur kernel, 10 iterations of CPCR run in 22ms, FFT without padding runs in 10ms and ``edgetaper'' padded FFT in 70ms.
Figure~\ref{fig:validations}
compares the running times of CPCR (run for 10 iterations) with padded/non-padded FFT for three image (resp. kernel) sizes: $500 \times 375$, $800 \times 800$ and $1280 \times 720$ ($27 \times 27$, $55 \times 55$ and $121 \times 121$) pixels.
Our method is marginally slower than FFT without padding in every configuration (within a margin of 20ms) but becomes much faster than FFT combined to ``edgetaper'' padding when the size of the kernel increases.
FFT with ``replicate'' padding runs in about the same time as FFT (no pad) and thus is not shown in Fig.~\ref{fig:validations}
.
The times have been averaged over 1000 runs. 

\textbf{Running times for computing the inverse kernels with Eq.~\eqref{eq:fftinversekernel}.}~Computing the inverse kernels $c_i$, with an ratio $w_{c}/w_{k}$ set to 2, takes $1.0\pm 0.2$ms for a blur kernel $k$ of size $27 \times 27$ and $5.4 \pm 0.5$ms (results averaged in 1000 runs) for a large $121 \times 121$ kernel.
Thus, the time for inverting blur kernels is negligible in the overall pipeline.

\textbf{CHQS validation.}~We compare several iterations of HQS using unpadded FFT (HQS-FFT (no pad.)), with ``replicate'' padding (HQS-FFT (rep. pad)), and the padding strategy proposed in \cite{kruse17learning} (HQS-FFT (FDN pad.)), CG (HQS-CG), or CPCR (CHQS) for solving 
the least-squares problem penalized with the TV-$\ell_1$ regularized in Eq.~\eqref{eq:penalizedproblem}
and use the same 160 blurry and noisy images than in previous paragraph as test set.
We set the number of HQS iterations $T$ to 10, run CPCR for 5 iterations and CG for at most 100 iterations.
We use $\lambda= 0.003$ and $\mu_t = 0.008 \times 4^t$ ($t = 0,\dots,T-1$). 
Table~\ref{tab:comparedtvl1model} compares the average PSNR scores obtained with the different HQS algorithms over the test set.
As expected, FDN padding greatly improves HQS-FFT results on larger kernels over naive ``replicate'' padding, \ie~``ker-4'' and ``ker-8'', but overall does not perform as well as CHQS.
For kernels 1, 2, 3 and 5, the four methods yield comparable results (within 0.1 dB of each other).
FFT-based methods are significantly worse on the other four, whereas our method gives better results than HQS-CG in general, but is 100 times faster.
This large speed-up is explained by the convolutional structure of CPCR whereas CG involves large matrix inversions and multiplications.
Figure~\ref{fig:qualitativetvl1model} shows a deblurring example from the test set.
HQS-FFT (with FDN padding strategy), even with the refined padding technique of \cite{kruse17learning}, produces a solution with boundary artifacts.
Both HQS-CG and CHQS restore the image with a limited amount of artifacts, but CHQS does it much faster than HQS-CG.
This is typical of our experiments in practice.

\begin{table}[t]
    \centering
    \adjustbox{max width=0.99\textwidth}{
    \begin{tabular}{lccccccc} \toprule
         & FCNN~\cite{zhang17fcnn} & EPLL~\cite{zoran11learning} & RGCD~\cite{gong2020deepgradient} & FDN~\cite{kruse17learning} & CHQS & $\text{LCHQS}_{\text{G}}$ & $\text{LCHQS}_{\text{F}}$ \\
        \midrule
        Levin~\cite{levin09understanding} & 33.08 & 34.82 & 33.73 & 35.09 & 32.12 & \underline{35.11 $\pm$ 0.05} & \textbf{35.15 $\pm$ 0.04}\\
        Sun~\cite{sun13edge} & 32.24 & 32.46 & 31.95 & 32.67 & 30.36 & \underline{32.83 $\pm$ 0.01} & \textbf{32.93 $\pm$ 0.01}\\
        \bottomrule
    \end{tabular}
    }
    \caption{PSNR scores for Levin~\cite{levin09understanding} and Sun~\cite{sun13edge} benchmarks, that respectively feature $0.5\%$ and $1\%$ noise. Best results are shown in bold, second-best underlined.
    The difference may not always be significant between FDN and LCHQS for the Levin dataset.}
    \label{tab:levin}
\end{table}

\begin{figure*}[t]
    \centering
    \adjustbox{max width=0.99\textwidth}{
    \begin{tabular}{cccc}
        \begin{subfigure}[b]{0.245\textwidth}\centering\includegraphics[scale=0.385]{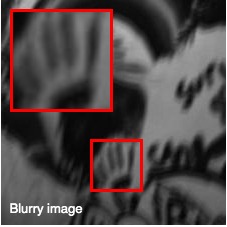}
        \end{subfigure} & 
        \begin{subfigure}[b]{0.245\textwidth}\centering\includegraphics[scale=0.385]{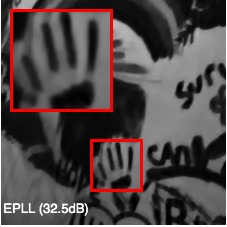}
        \end{subfigure} &
        \begin{subfigure}[b]{0.245\textwidth}\centering\includegraphics[scale=0.385]{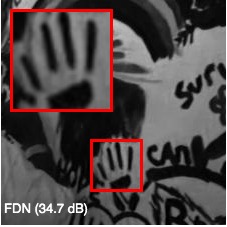}
        \end{subfigure} & 
        \begin{subfigure}[b]{0.245\textwidth}\centering\includegraphics[scale=0.385]{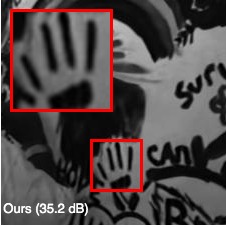}
        \end{subfigure} \\
    \end{tabular}
    }
    \caption{Comparison of state-of-the-art methods and the proposed LCHQS for one sample of the Levin dataset~\cite{levin09understanding} (better seen on a computer screen). FDN effectively removes the blur but introduces artifacts in flat areas, unlike EPLL and LCHQS.}
    \label{fig:levin}
\end{figure*}

\textbf{Discussion.}
These experiments show that CPCR always gives better results than CG in terms of PSNR, sometimes by a significant margin, and it is about 50 times faster. 
This suggests that CPCR may, more generally, be preferable to CG for linear least-squares problems when the linear operator is a convolution.
CPCR also dramatically benefits from its convolutional implementation on a GPU with speed similar to FFT and is even faster than FFT with FDN padding for large kernels.
These experiments also show that CHQS surpasses, in general, HQS-CG and HQS-FFT for deblurring.

Next, we further improve the accuracy of CHQS using supervised learning, as done in previous works blending within a single model variational methods and learning.

\subsection{Uniform deblurring}

We compare in this section CHQS and its learnable version LCHQS with the non-blind deblurring state of the art, including optimization-based and CNN-based algorithms.

\begin{figure}[t]
    \centering
    \adjustbox{max width=0.99\textwidth}{
    \begin{tabular}{cc}
        \begin{subfigure}[t]{0.49\textwidth}\centering\includegraphics[scale=0.335]{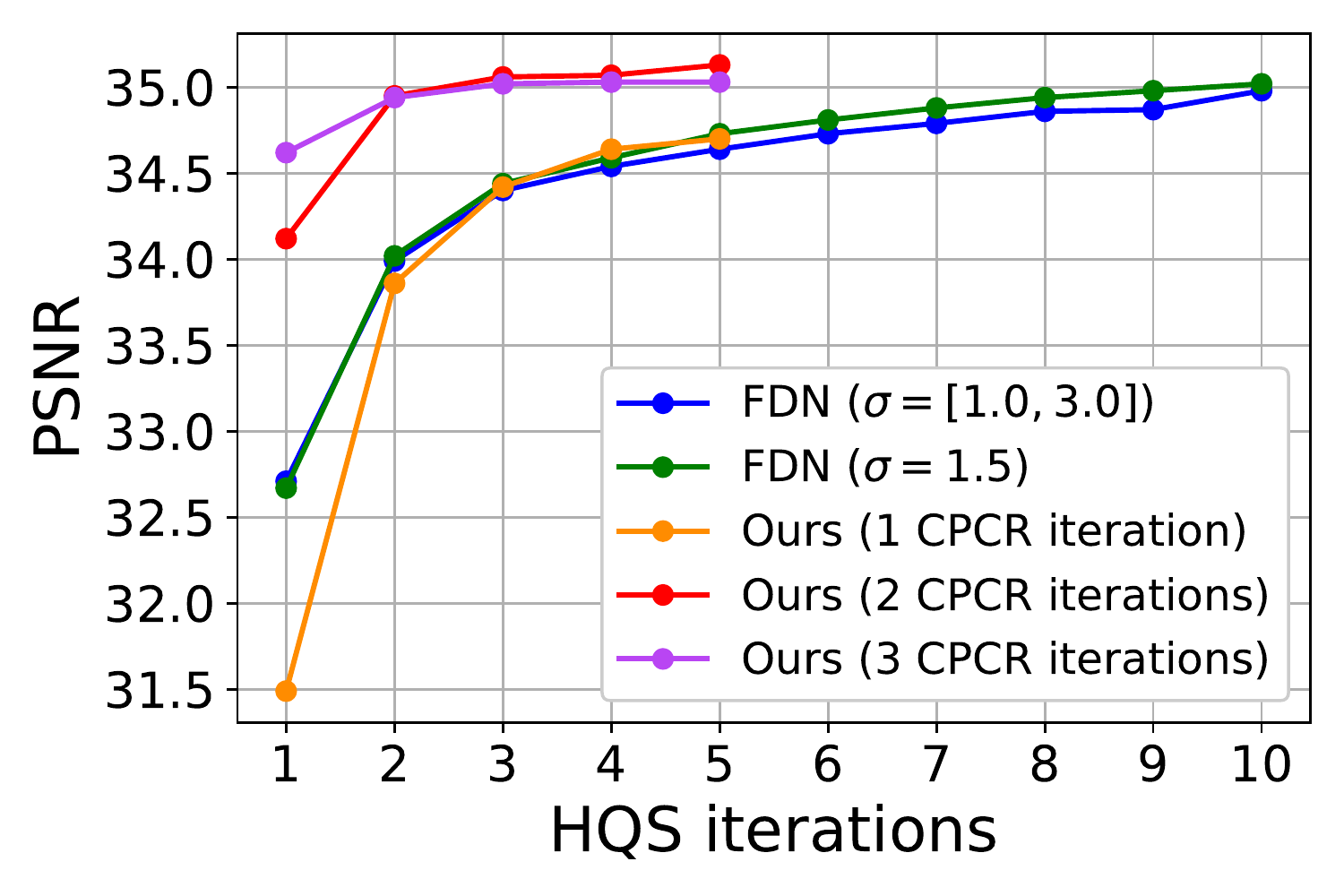}
        \end{subfigure} &  
        \begin{subfigure}[t]{0.49\textwidth}\centering\includegraphics[scale=0.335]{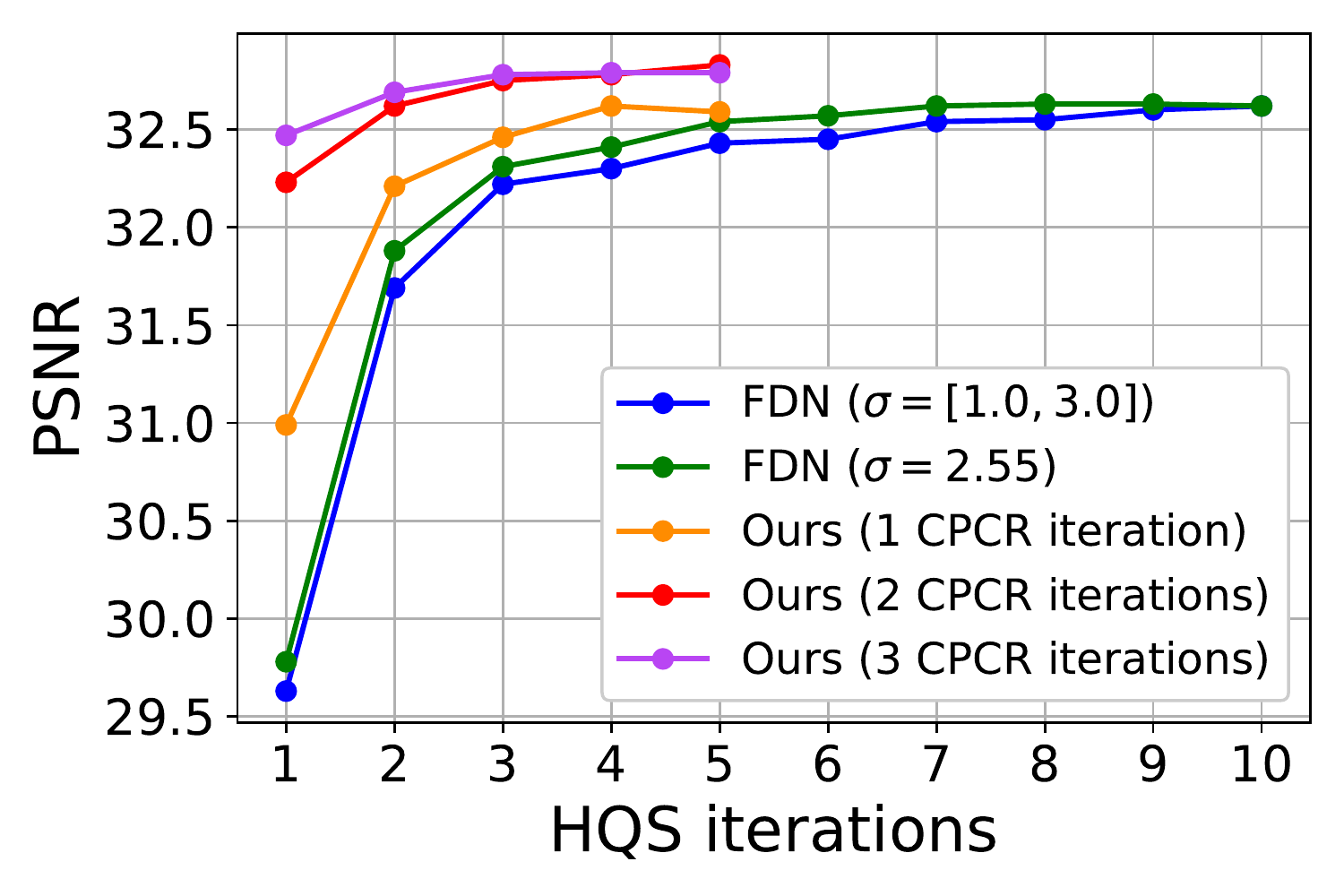}
        \end{subfigure}\\
    \end{tabular}
    }
    \caption{Performance of FDN \cite{kruse17learning} and LCHQS on the Levin~\cite{levin09understanding} (left) and Sun~\cite{sun13edge} (right) datasets.
    }
    \label{fig:comparehqsiterationslevinandsun}
\end{figure}

\textbf{Comparison on standard benchmarks.}~LCHQS is first trained by using the loss of Eq.~\eqref{eq:objectivefull} to supervise the output of each stage of the proposed model and second trained by only supervising the output of the final layer, in the manner of \cite{kruse17learning}.
The model trained in the first regime is named $\text{LCHQS}_{\text{G}}$ and the one further trained with the second regime is named $\text{LCHQS}_{\text{F}}$.
The other methods we compare our learnable model to are HQS algorithms solving a TV-$\ell_1$ problem: HQS-FFT (with the padding strategy of \cite{kruse17learning}), HQS-CG and CHQS, an HQS algorithm with a prior over patches (EPLL)~\cite{zoran11learning} and the state-of-the-art CNN-based deblurring methods FCNN~\cite{zhang17fcnn} and FDN~\cite{kruse17learning}.
We use the best model provided by the authors of \cite{kruse17learning}, denoted as $\text{FDN}_{\text{T}}^{\text{10}}$ in their paper.
Table~\ref{tab:levin} compares our method with these algorithms on two classical benchmarks. 
We use 5 HQS iterations and 2 CPCR iterations for CHQS and LCHQS.
Except for EPLL that takes about 40 seconds to restore an image of the Levin dataset~\cite{levin09understanding}, all methods restore a $255 \times 255$ black and white image in about 0.2 second. 
The dataset of Sun \textit{et al.} contains high-resolution images of size around $1000 \times 700$ pixels. EPLL removes the blur in 20 minutes on a CPU while the other methods, including ours, do it in about 1 second on a GPU.
In this case, our learnable method gives comparable results to FDN~\cite{kruse17learning}, outputs globally much sharper results than EPLL~\cite{zoran11learning} and is much faster.
As expected, non-trainable CHQS is well behind its learned competitors (Tab.~\ref{tab:levin}).

\textbf{Number of iterations for $\text{LCHQS}_{G}$ and CPCR.}~We investigate the influence of the number of HQS and CPCR iterations on the performance of $\text{LCHQS}_{G}$ on the benchmarks of Levin \textit{et al.}~\cite{levin09understanding} and Sun \textit{et al.}~\cite{sun13edge}.
FDN implements 10 HQS iterations parameterized with CNNs but operates in the Fourier domain.
Here, we compare $\text{LCHQS}_{G}$ to the FDN model trained in a stage-wise manner (denoted as $\text{FDN}_{\text{G}}^{\text{10}}$ in \cite{kruse17learning}).
Figure~\ref{fig:comparehqsiterationslevinandsun} plots the mean PSNR values for the datasets of Levin \textit{et al.} and Sun \textit{et al.} \cite{sun13edge} after each stage.
FDN comes in two versions: one trained on a single noise level (green line) and one trained on noise levels within a given interval (blue line).
We use up to 5 iterations of our learnable CHQS scheme, but it essentially converges after only 3 steps.
When the number of CPCR iterations is set to 1, FDN and our model achieve similar results for the same number of HQS iterations.
For 2/3 CPCR iterations, we do better than FDN for the same number of HQS iterations by a margin of +0.4/0.5dB on both benchmarks.
For 3 HQS iterations and more, LCHQS saturates but systematically achieves better results than 10 FDN iterations: +0.15dB for~\cite{levin09understanding} and +0.26dB for~\cite{sun13edge}.

\begin{figure*}[t]
    \centering
    \adjustbox{max width=0.99\textwidth}{
    \begin{tabular}{ccc}
        \begin{subfigure}[b]{0.33\textwidth}\centering\includegraphics[scale=0.238]{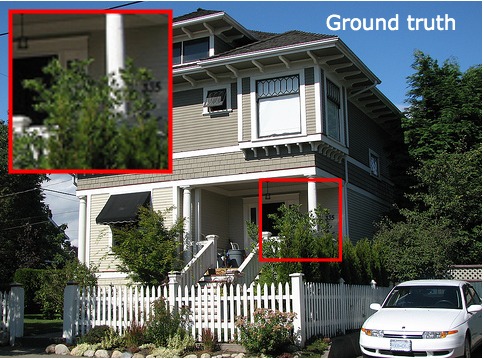}\end{subfigure} & 
        \begin{subfigure}[b]{0.33\textwidth}\centering\includegraphics[scale=0.238]{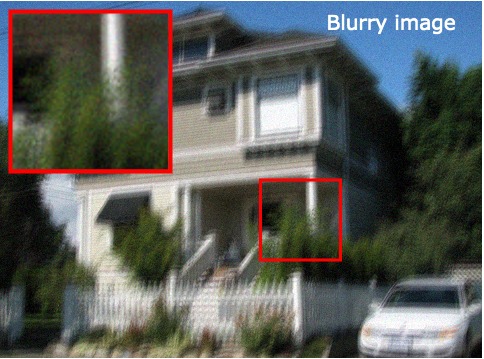}\end{subfigure} & 
        \begin{subfigure}[b]{0.33\textwidth}\centering\includegraphics[scale=0.238]{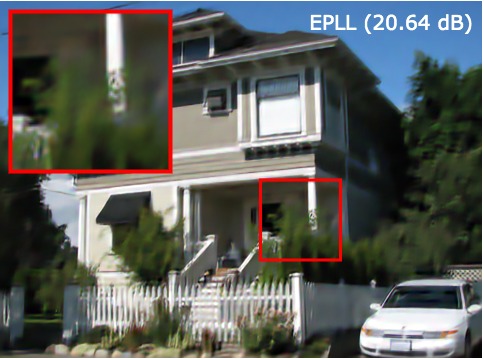}\end{subfigure} \\
        \begin{subfigure}[b]{0.33\textwidth}\centering\includegraphics[scale=0.238]{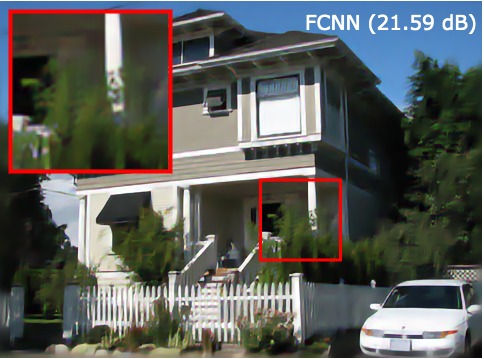}\end{subfigure} & 
        \begin{subfigure}[b]{0.33\textwidth}\centering\includegraphics[scale=0.238]{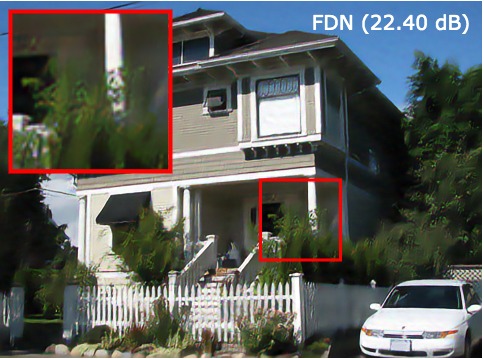}\end{subfigure} & 
        \begin{subfigure}[b]{0.33\textwidth}\centering\includegraphics[scale=0.238]{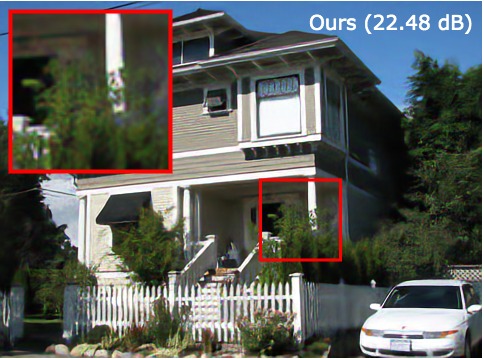}\end{subfigure} \\
    \end{tabular}
    }
    \caption{Example of image deblurring with an additive noise of $3\%$ (better seen on a computer screen). In this example, we obtain better PSNR scores than competitors and better visual results, for example details around the door or the leaves.}
    \label{fig:noisyexample}
\end{figure*}

\textbf{Robustness to noise.}~Table~\ref{tab:uniformpascalcropped} compares our methods for various noise levels on the 160 RGB images introduced previously, dubbed from ``PASCAL benchmark''.
FDN corresponds to the model called $\text{FDN}_{\text{T}}^{\text{10}}$ in \cite{kruse17learning}.
For this experiment, (L)CHQS uses 5 HQS iterations and 2 inner CPCR iterations.
We add $1\%$, $3\%$ and $5\%$ Gaussian noise to these images to obtain three different test sets with gradually stronger noise levels.
We train each model to deal with a specific noise level (non-blind setting) but also train a single model to handle multiple noise levels (blind setting) on images with 0.5 to 5$\%$ of white noise, as done in \cite{kruse17learning}.
For each level in the non-blind setting, we are marginally above or below FDN results.
In terms of average PSNR values, the margins are +0.12dB for $1\%$, +0.06dB for $3\%$  and -0.05dB for $5\%$ when comparing our models with FDN, but we are above the other competitors by margins between 0.3dB and 2dB.
Compared to its noise-dependent version, the network trained in the blind setting yields a loss of 0.2dB for 1$\%$ noise, but gains of 0.14 and 0.27dB for 3 and $5\%$ noises, showing its robustness and adaptability to various noises.
Figure \ref{fig:noisyexample} compares results obtained on a blurry image with $3\%$ noise.

\begin{table}[t!]
    \centering
    \begin{tabular}{lcccc} \toprule
        & $1\%$ noise & $3\%$ noise & $5\%$ noise &  Time (s) \\ \midrule
        HQS-FFT & 26.48 & 23.90 & 22.15 & 0.2 \\
        HQS-CG & 26.45 & 23.91 & 22.27 & 13 \\ 
        EPLL~\cite{zoran11learning} & 28.83 & 24.00 & 22.10 & 130 \\
        FCNN~\cite{zhang17fcnn} & 29.27 & 25.07 & 23.53 & 0.5\\
        FDN~\cite{kruse17learning} & \underline{29.42} & 25.53 & 23.97 & 0.6 \\
        \hdashline
        $\text{CHQS}$ & 27.08 & 23.33 & 22.38 & 0.3\\
        $\text{LCHQS}_\text{G}$ (non-blind) & \textbf{29.54 $\pm$ 0.02} & \underline{25.59 $\pm$ 0.03} & 23.87 $\pm$ 0.06 & 0.7\\
        $\text{LCHQS}_\text{F}$ (non-blind) & \textbf{29.53 $\pm$ 0.02} & 25.56 $\pm$ 0.03 & 23.95 $\pm$ 0.05 & 0.7\\
        $\text{LCHQS}_\text{G}$ (blind) & 29.22 $\pm$ 0.02 & 25.55 $\pm$ 0.03 & \underline{24.05 $\pm$ 0.02} & 0.7\\
        $\text{LCHQS}_\text{F}$ (blind) & 29.35 $\pm$ 0.01 & \textbf{25.71} $\pm$ \textbf{0.02} & \textbf{24.21} $\pm$ \textbf{0.01}  & 0.7\\
        \bottomrule
    \end{tabular}
    \caption{Uniform deblurring on 160 test images with $1\%$, $3\%$ and $5\%$ white noise. Running times are for an $500 \times 375$ RGB image. The mention ``blind'' (resp. ``non-blind) indicates that a single model handles the three (resp. a specific) noise level(s).}
    \label{tab:uniformpascalcropped}
\end{table}

\subsection{\textbf{Non-uniform motion blur removal}}

\begin{figure*}[t!]
    \centering
    \adjustbox{max width=0.99\textwidth}{
    \begin{tabular}{cccc}
        \begin{subfigure}[b]{0.245\textwidth}\centering\includegraphics[scale=0.18]{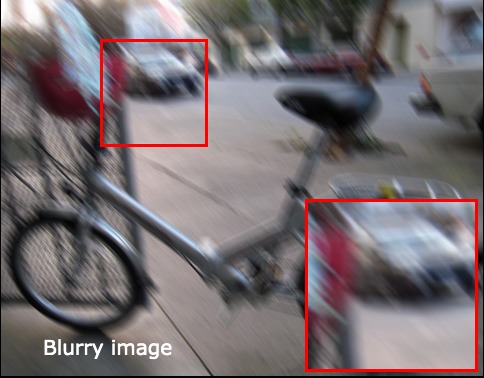}
        \end{subfigure} & 
        \begin{subfigure}[b]{0.245\textwidth}\centering\includegraphics[scale=0.18]{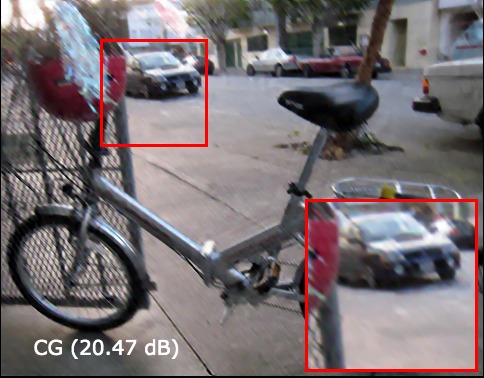}
        \end{subfigure} &
        \begin{subfigure}[b]{0.245\textwidth}\centering\includegraphics[scale=0.18]{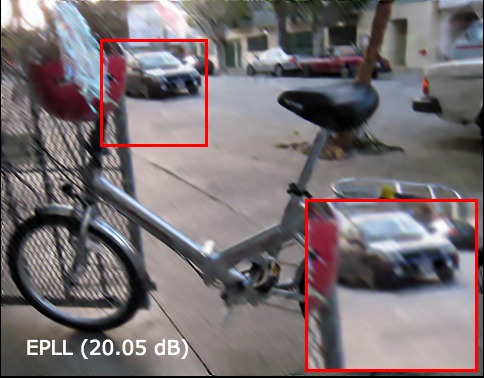}
        \end{subfigure} & 
        \begin{subfigure}[b]{0.245\textwidth}\centering\includegraphics[scale=0.18]{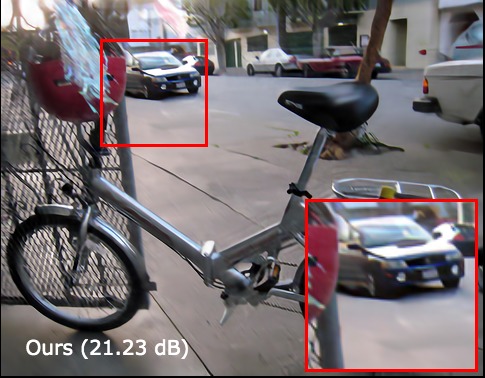}
        \end{subfigure} \\
    \end{tabular}
    }
    \caption{Non-uniform motion deblurring example with 1 $\%$ additive Gaussian noise (better seen on a computer screen). The car and the helmet are sharper with our method than in the images produced by our competitors.}
    \label{fig:nonuniformqualitative}
\end{figure*}

Typical non-uniform motion blur models assign to each pixel of a blurry image a local uniform kernel \cite{chakrabarti10analyzing}.
This is equivalent to replacing the uniform convolution in Eq.~\eqref{eq:penalizedproblem} by local convolutions for each overlapping patch in an image, as done by Sun \textit{et al.} \cite{sun15learning} when they adapt the solver of \cite{zoran11learning} to the non-uniform case.
Note that FDN~\cite{kruse17learning} and FCNN~\cite{zhang17fcnn} operate in the Fourier domain and thus cannot be easily adapted to non-uniform deblurring, unlike (L)CHQS operating in the spatial domain.
We handle non-uniform blur as follows to avoid computing different inverse filters at each pixel.
As in \cite{sun15learning}, we model a non-uniform motion field with {\em locally} linear motions that can well approximate complex {\em global} motions such as camera rotations.
We discretize the set of the linear motions by considering only those with translations (in pixels) in $\{1,3\dots,35\}$ and orientations in $\{0^\circ,6^\circ, \dots, 174^\circ \}$.
In this case, we know in advance all the 511 $35 \times 35$ local blur kernels and compute their approximate inverses ahead of time.
During inference, we simply determine which one best matches the local blur kernel and use its approximate inverse in CPCR.
This is a parallelizable operation on a GPU.
Table~\ref{tab:nonunifpascal} compares our approach (in non-blind setting) to existing methods for locally-linear blur removal on a test set of 100 images from PASCAL dataset non-uniformly blurred with the code of \cite{gong17motion} and with white noise.
For instance for $1\%$ noise, $\text{LCHQS}_\text{G}$ scores +0.99dB higher than CG-based method, and $\text{LCHQS}_\text{F}$ pushes the margin up to +1.13dB while being 200 times faster. 
Figure~\ref{fig:nonuniformqualitative} shows one non-uniform example from the test set.

\begin{table}[t]
    \center
    \adjustbox{max width=0.99\textwidth}{
    \begin{tabular}{lcccccc} \toprule
     & HQS-FFT & HQS-CG & EPLL~\cite{zoran11learning} & CHQS & $\text{LCHQS}_\text{G}$ & $\text{LCHQS}_\text{F}$ \\ \bottomrule
     $1\%$ noise & 23.49 & 25.84 & 25.49 & 25.11 & \underline{26.83 $\pm$ 0.08} & \textbf{26.98} $\pm$ \textbf{0.08} \\
     $3\%$ noise & 23.17 & 24.18 & 23.78 & 23.74 & \underline{24.91 $\pm$ 0.05} & \textbf{25.06} $\pm$ \textbf{0.06} \\
     $5\%$ noise & 22.44 & 23.10 & 23.34 & 22.65 & \underline{23.97 $\pm$ 0.05} & \textbf{24.14} $\pm$ \textbf{0.05} \\ 
      Time (s) & 13 & 212 & 420 & 0.8 & 0.9 & 0.9\\ \bottomrule
     \end{tabular}
     }
     \caption{Non-uniform deblurring on 100 test images with $1\%$, $3\%$ and $5\%$ white noise. Running times are for an $500 \times 375$ RGB image.}
     \label{tab:nonunifpascal}
\end{table}

\subsection{Deblurring with approximated blur kernels}

\begin{figure*}[t]
    \centering
    \adjustbox{max width=0.99\textwidth}{
    \begin{tabular}{cccc}
        \begin{subfigure}[b]{0.245\textwidth}\centering\includegraphics[scale=0.126]{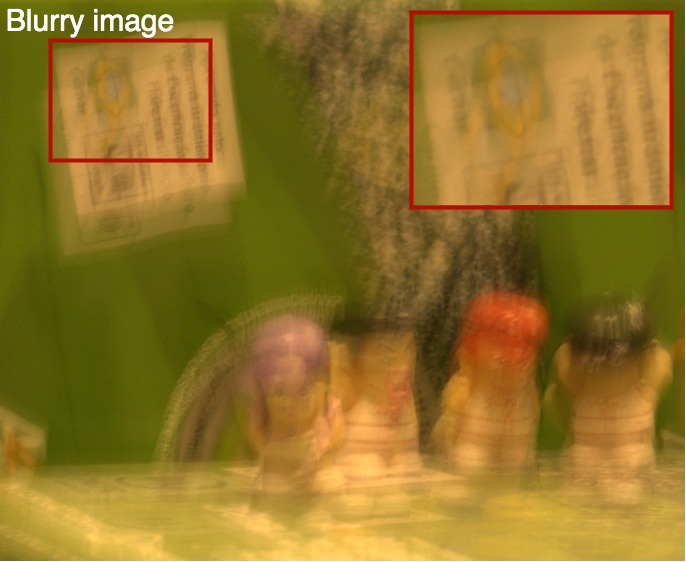}
        \end{subfigure} & 
        \begin{subfigure}[b]{0.245\textwidth}\centering\includegraphics[scale=0.126]{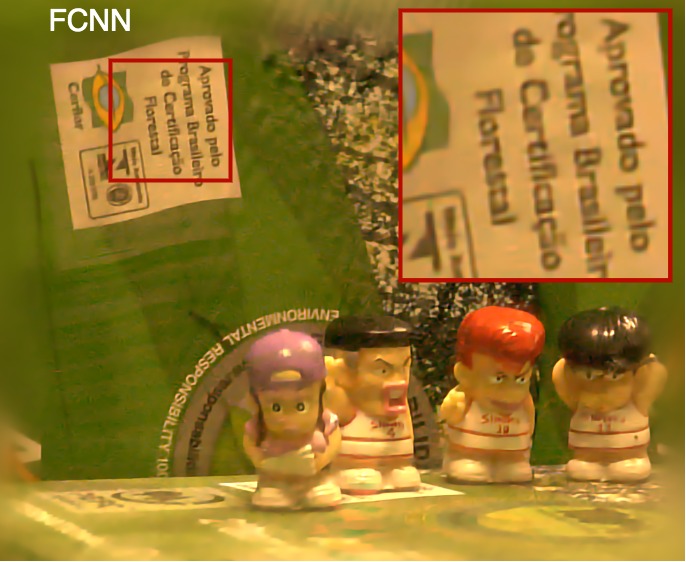}
        \end{subfigure} &
        \begin{subfigure}[b]{0.245\textwidth}\centering\includegraphics[scale=0.126]{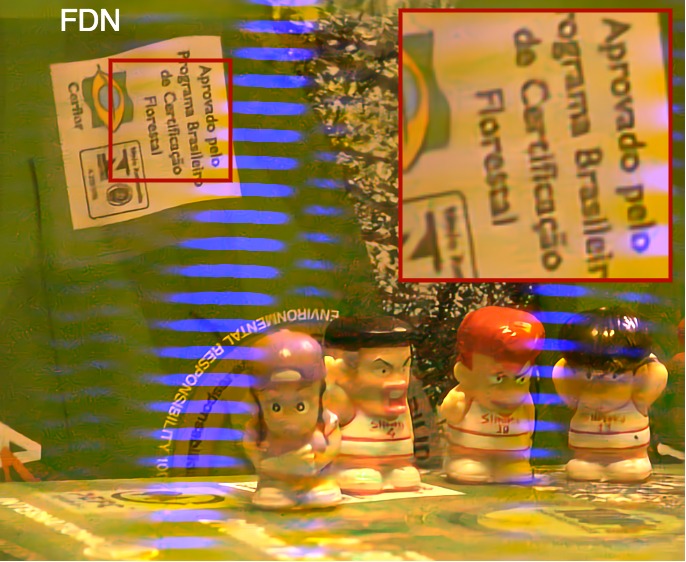}
        \end{subfigure} & 
        \begin{subfigure}[b]{0.245\textwidth}\centering\includegraphics[scale=0.126]{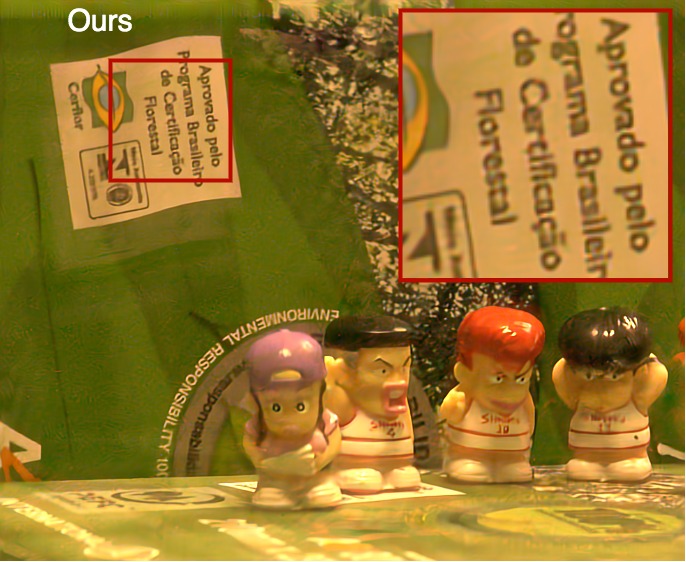}
        \end{subfigure} \\
    \end{tabular}
    }
    \caption{Real-world blurry images deblurred with an $101 \times 101$ blur kernel estimated with \cite{pan18blind}.
    We can restore fine details with approximate, large kernels.}
    \label{fig:realistic}
\end{figure*}

In practice one does not have the ground-truth blur kernel but instead an {\em approximate} version of it, obtained with methods such as \cite{pan18blind, whyte12nonuniform}.
We show that (L)CHQS works well for approximate and/or large filters, different from the ones used in the training set and without any training or fine-tuning. 
We show in Figure~\ref{fig:realistic} a deblurred image with an approximate kernel obtained with the code of \cite{pan18blind} and of support of size $101 \times 101$ pixels.
We obtain with LCHQS$_\text{F}$ (blind) of Table~\ref{tab:uniformpascalcropped} a sharper result than FCNN and do not introduce artifacts as FDN, showing the robustness of CPCR and its embedding in HQS to approximate blur kernels.
More results are shown in the supplemental material.

\section{Conclusion}

We have presented a new learnable solver for non-blind deblurring. 
It is based on the HQS algorithm for solving penalized least-squares problems but uses preconditioned iterative fixed-point iterations for the $x$-update.
Without learning, this approach is superior both in terms of speed and accuracy to classical solvers based on the Fourier transform and conjugate gradient descent.
When the preconditioner and the proximal operator are learned, we obtain results that are competitive with or better than the state of the art. Our method is easily extended to non-uniform deblurring, and it outperforms the state of the art by a significant margin in this case. We have also demonstrated its robustness to important amounts of white noise. Explicitly accounting for more realistic noise models \cite{foi08practical} and other degradations such as downsampling is left for future work.
\\

\noindent\textbf{Acknowledgments.}~This works was supported in part by the INRIA/NYU collaboration and the Louis Vuitton/ENS chair on artificial intelligence.
In addition, this work was funded in part by the French government under management of Agence Nationale de la Recherche as part of the ``Investissements d’avenir” program, reference ANR19-P3IA-0001 (PRAIRIE 3IA Institute).
Jian Sun was supported by NSFC under grant numbers 11971373 and U1811461.

%
%
\bibliographystyle{splncs04}
\bibliography{egbib}

\end{document}